\documentclass{article}
\usepackage{arxiv}

\usepackage{blindtext}
\usepackage{eso-pic}

\usepackage{cite}
\usepackage{amsmath,amssymb,amsfonts}
\usepackage{bookman}
\usepackage{amsmath,amssymb}
\usepackage{booktabs}
\usepackage{fancybox, graphicx}
\usepackage{algorithm}
\usepackage[noend]{algpseudocode}
\usepackage{textcomp}
\usepackage{xcolor}

\def\BibTeX{{\rm B\kern-.05em{\sc i\kern-.025em b}\kern-.08em
    T\kern-.1667em\lower.7ex\hbox{E}\kern-.125emX}}
\usepackage[draft]{hyperref}
\definecolor{darkred}{rgb}{0.85,0,0}
\definecolor{darkgreen}{rgb}{0,0.6,0}
\definecolor{darkblue}{rgb}{0,0,0.5}
\hypersetup{hidelinks, breaklinks, colorlinks,bookmarks=false,
  linkcolor={darkblue}, 
  citecolor={darkblue}, 
  urlcolor={darkblue}}

\usepackage{subfigure}
\begin{document}

\title{Neural Network based on Automatic Differentiation Transformation of Numeric Iterate-to-Fixedpoint}

\author{ 
	\href{https://orcid.org/0000-0001-9051-1370}{\includegraphics[scale=0.06]{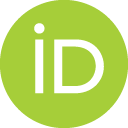}\hspace{1mm}Mansura Habiba} \\
	Dept of Computer Science\\
	Maynooth University\\
	Maynooth, Ireland \\
	\And
	\href{https://orcid.org/0000-0003-0521-4553}{\includegraphics[scale=0.06]{orcid.png}\hspace{1mm}Barak A. Pearlmutter} \\
	Department of Computer Science \& Hamilton Institute\\
	Maynooth University\\
	Maynooth, Ireland \\
	
}

\maketitle

\begin{abstract}
    This work proposes a Neural Network model that can control its depth using an iterate-to-fixed-point operator. The architecture starts with a standard layered Network but with added connections from current later to earlier layers, along with a gate to make them inactive under most circumstances. These ``temporal wormhole'' connections create a shortcut that allows the Neural Network to use the information available at deeper layers and re-do earlier computations with modulated inputs. End-to-end training is accomplished by using appropriate calculations for a numeric iterate-to-fixed-point operator. In a typical case, where the ``wormhole'' connections are inactive, this is inexpensive; but when they are active, the network takes a longer time to settle down, and the gradient calculation is also more laborious, with an effect similar to making the network deeper. In contrast to the existing skip-connection concept, this proposed technique enables information to flow up and down in the network. Furthermore, the flow of information follows a fashion that seems analogous to the afferent and efferent flow of information through layers of processing in the brain. We evaluate models that use this novel mechanism on different long-term dependency tasks. The results are competitive with other studies, showing that the proposed model contributes significantly to overcoming traditional deep learning models' vanishing gradient descent problem. At the same time, the training time is significantly reduced, as the ``easy'' input cases are processed more quickly than ``difficult'' ones.
\end{abstract}

\section{Introduction}
\label{sec:ch9-intro}

Traditional Memory augmented neural networks (MANN) \cite{santoro2016one} as well as Recurrent Neural network (RNN) preserve previous states in the model by explicitly storing previous hidden state in the memory. However, if storing states to memory is too frequent, that can cause the memory to become very unstable. At the same time, the gradient at the beginning of the training is not stable, and a large fraction of memory gets overwritten at each step during the training. Unstable memory causes fast vanishing of memory and gradients.

Another well-known technique to preserve memory is Wormhole connection \cite{gulcehre2017memory}.  Wormhole connection in MANN such as neural Turing machines \cite{ntm} provides a shortcut connection to the previous hidden state through time by explicitly storing the previous hidden state in the memory. Temporal Automatic Relation Discovery in Sequences (TARDIS) \cite{gulcehre2017memory} shows that the Wormhole connection created by the controller of the MANN can significantly reduce the effects of the vanishing gradients. Wormhole connection for MANN shortens the paths that the signal needs to travel between the dependencies. At every step of training, the controller $\phi$ in TARDIS as shown in Eq.~\eqref{eq:tardis_controller}, controls the hidden state $\mathbf{h}_{t-1}$ based on the content of weights $\mathbf{w}_{t}$ and memory $\mathbf{M}_{t}$ read from external memory.

\begin{equation}
\mathbf{h}_{t}=\phi(\mathbf{x}_{t}, \mathbf{h}_{t-1},( \mathbf{M}_{t}^{\top} \mathbf{w}_{t}^{r}))
\label{eq:tardis_controller}
\end{equation}


On the other hand, a deep neural network, such as a Residual neural network, creates multiple layers (from 50 to 152 or more) of repetitive blocks. A ResNet with 100 layers provides 94\% accuracy for the MNIST dataset. A recent work suggested 1001-layer deep ResNet architecture \cite{he2016identity}. This 1001-layer deep ResNet generalizes data modelling better than the original ResNet-152-layer. This model has a 10-fold(10x) number of layers that improve the results where the original ResNet start to overfit. However, depth is not the only strength of this 1001-layer deep ResNet; identity mapping and better technique to resolve vanishing gradient problem helped a lot to achieve a better result.  Making a deep neural network is not always effective. Therefore, recent research works are focusing on Wide Residual Neural network \cite{he2016identity} and U-Net \cite{ronneberger2015u}.

In this work, I introduce a novel neural network model that uses AD Transformation of Numeric Iterate-to-Fixedpoint methodology to train each batch in the training dataset. In addition to dynamic training time for each layer, the proposed model can push backwards during training using a novel block in the architecture of a deep neural network. The main contributions of this work are as follows.

\begin{itemize}
	\item A novel deep neural network that leverage Fixed point iteration for batch training which is also controlled by both forward mode as well as backward mode of Algorithmic differentiation (AD).
	\item New technique for training a deep neural network with the dynamic amount of time for different layer by creating push-backwards connection with the previous layer of the model.
\end{itemize}

\section{Background}
The core concept of the proposed new model is to design a deep network that can have lower depth but higher training time. Instead of a fixed training time on each layer, this new model can dynamically change the training time for each layer. Therefore, the target model can have a small depth level, i.e. 50, but each layer can have a dynamic training time. Mainly, this new model trains each layer until it achieves an error less than or equal to a threshold value for error that I call tolerance error ($\varepsilon$).  For example, instead of using a 256-layer ResNet, we can achieve similar accuracy by learning the data using a 50-layer ResNet for a longer time.  
It is well-known that a machine learning algorithm can reach the final outcome sometimes right away, or sometimes it needs to think for a longer time. 

\begin{equation}
y=f(x)
\label{eq:net}
\end{equation}

For each individual input $x$ used in a deep learning model $f$ as shown in Eq.~\eqref{eq:net}, the computation time of primal solution $y$ varies. For some input $x_a$, the computation time is short, and the corresponding solution can be achieved in time $T1$. Similarly, for some input  $x_b$, the computation time, $T2$, for solution can be significantly higher, where $T2 >> T1$. As a result, a constant depth for the deep learning neural network model does not help to optimize the training. In addition, the depth of the model can vary for different batch input during the training. Moreover, the memory requirement for different model usually either constant such as Neural ODE \cite{chen2018neural} or constantly increasing overtime throughout the training period. Most deep neural network adopts push-forward technique where the training is uni-directional and skips connections are created to a future layer instead of a previous layer. However, in reality, learning a certain layer for a long time can achieve better accuracy, therefore. It is essential to create a connection with the previous layer by adopting a push-backward technique.  

For a traditional ResNet, each layer is executed in a sequence, or the IdentityBlock create a skip connection between the Basic ResNet Block of the current layer and the last ReLU layer. ResNet model always moves forward.  Identity Block help to optimize the training for ResNet, but at the same time, it also possesses some limitation. Creating skip connection using Identity Block enable the gradient to avoid the main-stream flow of residual block weights, and it can avoid learning anything during training. However, this technique also provides a hypothesis that it is possible that only a few blocks can represent the hidden dynamics of the dataset while many other blocks contain very little information to learn useful representations. Therefore, learning only a few blocks for a long time can optimize the performance of deep neural network models. In this work, I introduce Fixed-point iteration \cite{pearlmutteralgorithmic} to control the learning or training duration for each block. Instead of learning each block, this new model focuses on blocks that provide useful representations for the hidden dynamics of the data for a longer time period.

\section{Model Design}

A ResNet usually have three different blocks
\begin{enumerate}
    \item  basic - two consecutive 3 × 3 convolutions followed by a  batch normalization and ReLU unit
    \item bottleneck - one  3 × 3 convolution surrounded by dimensionality reducing and expanding 1×1 convolution layers
    \item identityBlock - one skip connection with input and the ReLU unit 
\end{enumerate}

\begin{figure}[htb]
	\centering
	\includegraphics[width=0.69\textwidth]{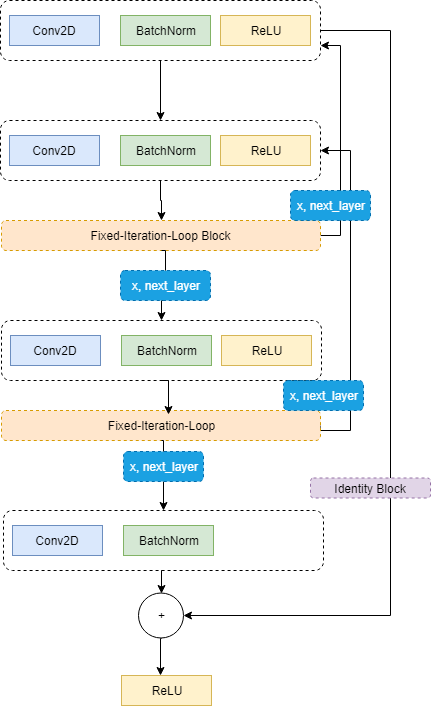}
	\caption{Block Diagram of architecture of push-backward technique for deep model}
	\label{fig:res-net-fp}
\end{figure}

The $Fixed-Point-Iterator$ block in Fig.~\ref{fig:res-net-fp} takes the input $(x)$ and run a fixed-point iteration loop until the error is less than or equal to the threshold for tolerance ($\varepsilon$). The output for $Fixed-Point-Iterator$ block is the output (x) and the next selected layer to be trained $\mathcal{L}$. If ($\mathcal{L}$) refers to a previous layer, the training restarted from that previous layer by creating a push-backward connection between the selected layer ($\mathcal{L}$) and the main-stream for training. As shown in Fig~\ref{fig:res-net-fp}, $Fixed-Point-Iterator$ block can send the training to the previous layer or can forward it to the next layer of a Residual neural network. 

At every layer, a local objective function finds the \emph{fixed point}.  Eq.~\eqref{eq:fixed-loop} shows the condition for the \emph{fixed point}. This loop iterate until difference between consecutive values $x_{t-1}$ and $x_{t}$ of variable $x$ is less than the tolerance ($\varepsilon$) or the number of iteration is more than max-iter (the maximum number of iteration for each \emph{Fixed-Point-Iterator} block).  $\mathcal{F}$ the actual numeric fixed-point, $z = x_{\infty}$, assuming that the objective function $g(\hat{A}\cdot b)$ has appropriate convergence properties.  $\mathcal{F}$ continues looking for $z$ until $\left\|\mathrm{x}_{t}-\mathrm{x}_{t-1}\right\|$ is less than some predefined threshold $\varepsilon$. This loop iterate until difference between consecutive values $x_{t-1} and x{t} $ of variable $x$ is less than the tolerance ($\varepsilon$).

\clearpage
\begin{equation}
\begin{array}{l}
\text{for $t=1, \ldots, \infty$ \{} \\
\quad \mathrm{x}_{t} \leftarrow g\left(\mathrm{x}_t, \mathrm{\alpha}, H_{t-1},O_{t-1}\right)\text{;} \\
\quad \text{if $\left\|\mathrm{x}_{t}-\mathrm{x}_{t-1}\right\| \leq \varepsilon$ break; \}} \\
\mathrm{z}=\mathrm{x}_{t}
\end{array}
\label{eq:fixed-loop}
\end{equation}

\begin{figure}[htb]
    \centering
    \includegraphics[width=\columnwidth]{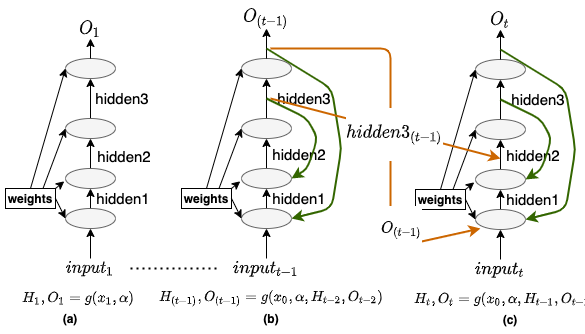}
    \caption{The workflow for objective Neural Network $g$ in Eq.~\eqref{eq:fixed-loop}}
    \label{fig:g}
\end{figure}

Here $g$ computes the cost for each step locally. Fig.~\ref{fig:g}(c) shows that $g$ takes the input of any current state($x_{t}$), weight($\alpha$), hidden matrix($H_{t-1}$) and output ($O_{t-1}$) of previous step($t-1$) as input. For first iteration,t=1, $H_0=null$ and $O_0 =null$, therefore, $H_1, O_1= g(x_1, \alpha)$. Here, the hidden matrix $H1={hidden1, hidden2, hidden3}$ as shown in Fig.~\ref{fig:g}(a). The hidden matrix of previous step(t-1) is used as input for  $g$ in current step (t). Fig.~\ref{fig:g}(b) shows g computes the output for previous step (t-1), which is used as input for current step t as shown as shown in Fig.~\ref{fig:g}(c). The  loop $t=1, \ldots, \infty$ iterates until $z$ settle down. However, with the "wormhole" reverse connections as shown in "green" color in  Fig.~\ref{fig:g}, the execution of $g$ may iterate a bunch of times for $z$ to settle down.

Fig.~\ref{fig:res-net-fp} shows that the sequence of three blocks repeat for a ResNet model based on the number of layer. These repetitive network can be described as Eq.~\eqref{eq:resnet-ch9}. 

\begin{equation}
   x_D=\sum_{i=d}^{D-1}\mathcal{F}(x_i, W_i)
   \label{eq:resnet-ch9}
\end{equation}

Here, $\mathcal{F}$ is a residual function. Between the shallow layer (d) and deep layer (D), some layer $(Dw)$ contributes to the final result more than other layers $(Dp)$. If a ResNet model is of depth D with number of layers= D, Eq.~\eqref{eq:depth} shows the distribution of layers of ResNet model.

\begin{equation}
   D = \sum_{j=0}^{P}Dw + \sum_{j=0}^{K}Dp  
   \label{eq:depth}
\end{equation}

Here, lets consider  layers denoted by $Dw$, learn useful representations of the hidden dynamics of the system, where layers in ResNet model architecture, denoted by $Dp$, provide very little information with small contribution to the final goal for learning the system. To optimize the performance of the model, it is essential to identify layers ($Dw$) with useful representations of the hidden dynamics. In this work, I introduce a new model that can identify the $Dw$ layers and can learn the hidden dynamics of the system by learning these layers only. This new model has a $fixed-point-iterative$ operator to control the training for the  ($Dw$) layers. The $fixed-point-iterative$ operator continue training for each  ($Dw$) layer until one of the following two conditions are met:

\begin{itemize}
    \item The number of iteration for each block training exceeds the $max_iter$
    \item the corresponding training loss is less than tolerance error ($\varepsilon$).
\end{itemize}

This process a new block called $Fixed-Point-Iterator$ block in the architecture of ResNet Model as shown in Fig.~\ref{fig:res-net-fp}. $Fixed-Point-Iterator$ block trains current $Dw$ layer.  Once the training for current $Dw$ layer is complete in the, $Fixed-Point-Iterator$ block identified the next layer $\mathcal{L}$ and create a connection between  $\mathcal{L}$ layer and the main stream of training. $Fixed-Point-Iterator$ block uses $fixed-point-iterative$ operator to control the training for current $Dw$ layer and enforce the above mentioned conditions to stop the loop. The $fixed-point-iterative$ operator leverage Banach Fixedpoint finder \textbf{\textit{fix}} :$(\beta \rightarrow \beta) \rightarrow \beta$ that takes a contraction of a closed region which contains the initial point. The reverse AD transform has signature as shown in  Eq.~\eqref{eq:ad_reverse}.

\begin{equation}\begin{array}{cc}
\overleftarrow{\mathcal{J}}:\left(\alpha_{1} \rightarrow \cdots \rightarrow \alpha_{n} \rightarrow \beta\right) \rightarrow\\
\alpha_{1} \rightarrow \cdots \; \rightarrow \alpha_{n} \rightarrow\\
\left(\beta \times\left(T^{*} \beta \rightarrow\left(T^{*} \alpha_{1} \times \cdots \times T^{*} \alpha_{n}\right)\right)\right)
\label{eq:ad_reverse}
\end{array}\end{equation}

The forward and reverse transform of \textbf{\textit{fix}} are shown as \eqref{eq:fix}.

\begin{equation}
z=\operatorname{fix}(f x) \quad f: \alpha \rightarrow \beta \rightarrow \beta
\label{eq:fix}
\end{equation}

\begin{equation}\begin{array}{cc}
T z=\operatorname{fix}(\overrightarrow{\mathcal{J}} f(T x)) & \bar{f}: T^{*} \beta \rightarrow\left(T^{*} \alpha \times T^{*} \beta\right) \\
T^{*} x=\rho_{1}\left(\operatorname{fix}(\underbrace{\rho_{2}(\overleftarrow{\mathcal{J}} f x z)}_{\bar{f}} \circ\left(+\left(T^{*} z\right)\right) \circ \rho_{2})\right)
\label{eq:fix2}
\end{array}\end{equation}

\begin{figure}[ht]
	\centering
	\includegraphics[width=0.45\textwidth]{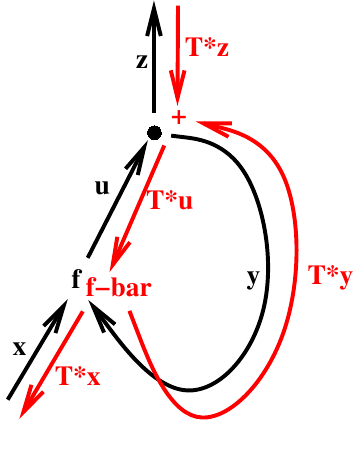}
	\caption{Forward and backward computation of proposed model}
	\label{fig:fixedpoint-reverse-graph}
\end{figure}

Here in Eq.~\eqref{eq:fix}, $f$, is a Neural Network. The weight to network $(f)$ is defined by $\alpha$. $\beta$ represents the activity of all input. Therefore, $(\beta \rightarrow \beta) \rightarrow \beta$ can be used to compute the next moment activity. Table \ref{tab:notation} describes different parameters in Equations \ref{eq:ad_reverse}, \ref{eq:fix} and \ref{eq:fix2} . Fig~\ref{fig:fixedpoint-reverse-graph} explains the computational graph for a single computational block of  proposed TWNN based on Eq.~\eqref{eq:ad_reverse}.

The gradient for each fixed-point iteration loop for each layer are also calculated over same number of iteration in a loop.


\begin{table}[htb]
	\centering
	\caption{The Notation Used in Temporal Wormhole neural Network}
	\label{tab:notation}

	\begin{tabular}{ll}
		\toprule
		$\overleftarrow{\mathcal{J}}$& Forward mode Algorithmic differentiation(AD) of Network $f$ \\
		$\overrightarrow{\mathcal{J}}$& backward mode Algorithmic differentiation(AD) of Network $f$  \\
		$\alpha$& Weight to the proposed network  \\
		$\beta$& Activity to the proposed network  \\
		$f$& Neural  Network  \\
		$\bar{f}$& Neural  Network for efferent connection\\
		$T^{*}z$& Gradient  of all the activity with respect to z  \\
		$T^{*}\alpha$& Gradient  of all the activity with respect to $\alpha$ \\
		$T^{*}\beta$& Gradient of all the activity with respect to $\beta$  \\
		$T^{*}x$& Gradient of all the activity with respect to $x$  \\
	\end{tabular}

\end{table}

\begin{figure}[ht]
  \centering
  \includegraphics[width=0.75\textwidth]{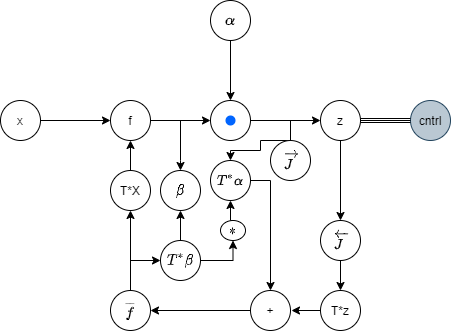}
  \caption{Loop for Fixed-Point iteration of the model training method}
  \label{fig:loop}
\end{figure}

As shown in Eq.\eqref{eq:fix}, the proposed Temporal Wormhole Neural Network (TWNN) uses Fixed-point iterations \cite{pearlmutteralgorithmic}. For each input $x$, the forward pass of TWNN continue iterate until the output $z$ in Eq.\eqref{eq:fix} settles. Fig \ref{fig:loop} shows the loop for each fixed-point iteration for each batch input $x$. At the end of each loop, the controller $cntrl$ checks either z settles by checking if the conditions in Eq~\eqref{eq:fixed-loop} are met.

Algorithm~\ref{alg:TemporalWormhole} explains the training method for proposed TWNN, the $model$ in algorithm~\ref{alg:TemporalWormhole} can be any Neural Network block such as GRU \cite{dey2017gate}, ResNet \cite{keeler1994residual} and others. 
\begin{algorithm}
	\caption{Fixed point temporal wormhole based Neural Network}
	\hspace*{\algorithmicindent} \textbf{Input:} Inputs are divided among batches, for each loop, a batch of inputs $x^{0}_{b}$, corresponding target $y_{b}, $ learning rate $l_{r}$, tolerance error ($\varepsilon$), maximum iteration $max_itr$;    \\
	\hspace*{\algorithmicindent} \textbf{Output:} predicted output z
	\begin{algorithmic}[1]
		\Procedure{$trainTWNNModel$}{$ x\_init$}
		\State  $ model, \alpha \leftarrow initializeNNBlock()$
		\For{ each $b$ in range $(total\_batch)$}
		\State $model_block \leftarrow next(model\_layers)$
		\State $loss = 1.0$
		    \While{ $model_block != null$}
			        \State  $ z_{b} \leftarrow Fixed-Point-Iterator(model_block, \alpha, x_{b} )$
			        \State $ loss = LossFun(z, y_{b})$
			    \State $model_block = next(model\_layers) || identify\_prev\_layer(model) $
			\EndWhile
		\EndFor
		\Return $ trained_model	$ \\
		\EndProcedure

	\end{algorithmic}
	\label{alg:TemporalWormhole}
\end{algorithm}

Proposed TWNN is initialized with the initial parameter $params$. \emph{Fixed-Point-Iterator} can be computed using Forward mode AD and backward mode AD as shown in Eq.~\eqref{eq:fp}.

\begin{equation}
	\textit{fixed-point-iterative} = (\overrightarrow{\mathcal{J}}\{f\},\overleftarrow{\mathcal{J}}\{f\})
\label{eq:fp}
\end{equation}

\begin{figure*}[htb]
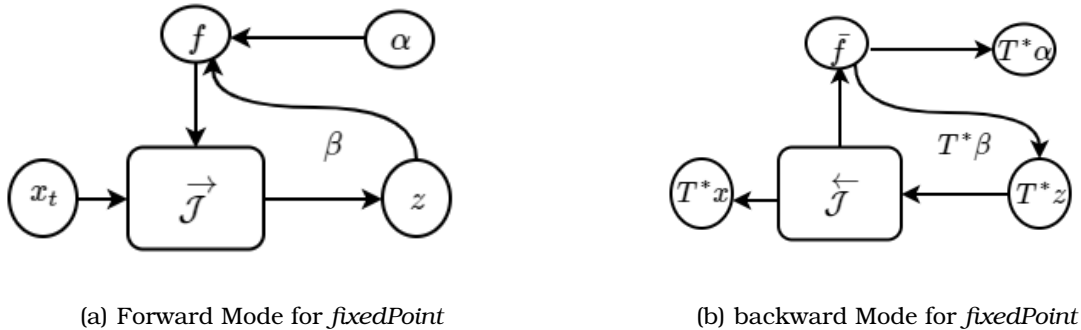

	\hfill
	\subfigure[Forward Mode for \emph{fixedPoint}]{\includegraphics[width=0.45\textwidth]{./ch9/images/fixed_point}}
	\hfill
	\subfigure[backward Mode for \emph{fixedPoint}]{\includegraphics[width=0.45\textwidth]{./ch9/images/fixed_point_bw}}
	\caption{Forward and backward computation of the controller of TWNN}
	\label{fig:fixedPoint}
\end{figure*}

Algorithm \ref{alg:fix} describes the forward pass and backward pass for a single block of TWNN. Figure \ref{fig:fixedPoint} explains the forward and backward computation of the \emph{fixed-point-iterative} operator of proposed TWNN.

\begin{algorithm}
	\caption{Forward and Backward Pass for fix function for a single block of TWNN}
	\hspace*{\algorithmicindent} \textbf{Input:} The value of x at time t $x_t$, parameters $\alpha$;    \\
	\hspace*{\algorithmicindent} \textbf{Output:} Gradient of x
	\begin{algorithmic}[1]
	\Procedure {$fix$} {$f, \alpha, x_{t}$}
	\State \textbf{Forward Pass:}
	\State  $ z, \beta \leftarrow f (x_{t}, \alpha)$
	\State  $ T^{*}z , T^{*}\alpha = grad(f, z,  \beta)$
	\State \textbf{Backward Pass:}
	\State  $ T^{*}x, T^{*}\beta \leftarrow \overline{f} (T^{*}z + T^{*}\beta)$ \\
	\Return $ T^{*}x $
	\EndProcedure

\end{algorithmic}
\label{alg:fix}
\end{algorithm}

\section{Experimental Result}
The performance of proposed model is evaluated against two common tasks:
\begin{enumerate}
    \item MNIST \cite{mnist} classification and
    \item Sine wave generation.
\end{enumerate}
 
 The training parameters for the two tasks are described in Table \ref{tab:parameters}.
 
\begin{table}[ht!]
	\centering
	\caption{Parameters for proposed model evaluation for MNIST classification}
	\label{tab:parameters}

	\begin{tabular}{ll}
		\toprule
		Learning Rate ($l_{r}$)& 0.001 \\
		Tolerance Error ($\varepsilon$)& 0.001  \\
		Batch Length Size &64 \\
		Maximum Number of Iteration per batch ($max\_iter$)& 300  \\
        $\varepsilon$ & 0.01\\
        \bottomrule
	\end{tabular}

\end{table}

\subsection{Task-A : CIFR  classification }

\begin{figure} [htb]
    \centering
    \includegraphics[width=\columnwidth]{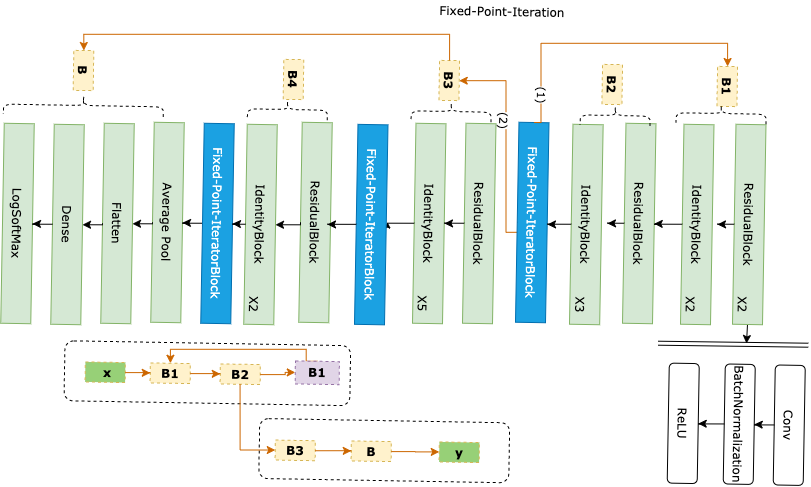}
    \caption{TWNN model for CIFR training}
    \label{fig:fpnet}
\end{figure}

For the training for this task, we have chosen a TWNN model described in Fig. ~\ref{fig:fpnet}. After the second block and fourth block, there is \emph{Fixed-Point-iterator} Block. After finishing 2nd block of the ResNet model, \emph{Fixed-Point-iterator} Block identify the next suitable block as B1 and the model continue training B1 and B2 blocks until the $\text{loss} > \varepsilon$  or $\text{iteration} < \text{max-iter}$. After the condition is met,  \emph{Fixed-Point-iterator} forward the training to the B3 layer. Then, \emph{Fixed-Point-iterator} Block after B3 identify the next layer as the final layer of the network and forward the training to layer B. 
Instead of training a deep ResNet-50 layer for this training, we trained a less deep RestNet-36 model. However, we trained specific layers of the model for a longer time.  Therefore, the memory consumption is significantly less than DNN models. This model skipped block B4 during training. Therefore, the training time is not higher than ResNet. 

Moreover, they show that the TWNN offers better results than the different DNN configurations. Table \ref{tab:result} shows the difference between the test accuracy of different similar configurations of ResNet and TWNN for CIFR.

\begin{table}[]
    \centering
    \caption{Comparative accuracy for proposed model against ResNet }
    \begin{tabular}{c|c}
        Neural Network & Accuracy \\
        \toprule
        RestNet- 36 & 88.25\% \\
        RestNet- 50 & 90.63\% \\
        TWRNN & 96.54\% \\
        \bottomrule
    \end{tabular}
    
    \label{tab:result}
\end{table}

\subsection{Task-B: Sine Wave Generation}
In this task, a corresponding sine wave is generated using the proposed TWNN model for a time series T.  Table \ref{tab:parameters-taskb} shows the parameters used in this task.
\begin{table}[ht!]
	\centering
	\caption{Parameters for proposed model evaluation for CIFR classification}
	\label{tab:parameters-taskb}

	\begin{tabular}{ll}
		\toprule
		Learning Rate ($l_{r}$)& 0.001 \\
		Tolerance Error ($\varepsilon$)& 0.001  \\
		Sequence Length Size &10000  \\
		Maximum Number of Iteration per batch ($max\_iter$)& 300  \\

	\end{tabular}
\end{table}

For this task, we take a simple Neural Network as the objective function shown in \ref{alg:fixet}. $loss$ function as shown in \ref{alg:net} used this $net$ procedure to compute gradient for the proposed TWNN.

\begin{algorithm}
	\caption{Neural Network as the objective function for the proposed model}

	\hspace*{\algorithmicindent} \textbf{Input:} The value of x at time t $x_t$, parameters $\alpha$;    \\
	\hspace*{\algorithmicindent} \textbf{Output:} the predicted value at time t  $x_t$
\begin{algorithmic}[1]
	\Procedure {$net$} {$f, \alpha, x_{t}$}
	\State  $ w1, b1, w2, b2 \leftarrow \alpha$
	\State  $	hidden \leftarrow tanh(dot(w1, x_{t}) +b1 )$
	\State  $y_{t }\leftarrow  sigmoid(dot(w2, hidden)+b2)$\\
   	\Return  $y_{t}$
   	\EndProcedure
\end{algorithmic}
\label{alg:fixet}
\end{algorithm}

\begin{algorithm}
	\caption{Loss function for the proposed model}

	\hspace*{\algorithmicindent} \textbf{Input:} The value of x at time t $x_t$, parameters $\alpha$, target ;    \\
	\hspace*{\algorithmicindent} \textbf{Output:} the predicted value at time t  $x_t$
\begin{algorithmic}[1]
	\Procedure {$loss$} {$net, \alpha, inputs, target$}
	\State  $ solver \leftarrow fix(net, \alpha, inputs)$
	\State  $	predictions \leftarrow solver(inputs, params)$    \\
   	\Return  $mean((targets - predictions)**2)$
   	\EndProcedure
\end{algorithmic}
\label{alg:net}
\end{algorithm}

\begin{figure*}[htb]
	\hfill
	\subfigure[Loss for 300 iterations]{\includegraphics[width=0.45\textwidth]{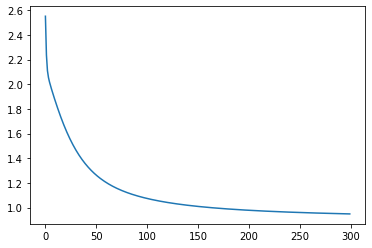}}
	\hfill
	\subfigure[Generated Sine Wave for time series]{\includegraphics[width=0.45\textwidth]{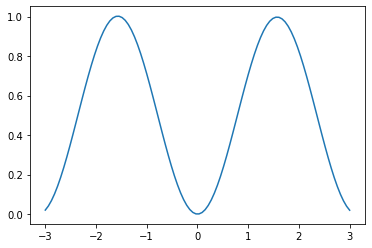}}
	\caption{Result of continuous Sine wave generation by TWNN}
	\label{fig:result-sine}
\end{figure*}

Fig.~\ref{fig:result-sine}(a) shows the gradients are optimized within 300th iteration and loss is stable. Also Fig.~\ref{fig:result-sine}(b) shows the generated sine wave by training TWNN model for 300 iterations .

\subsection{Discussion}
 The proposed TWNN model can be a suitable alternative against  deep ResNet models with larger batch sizes; the TWNN model can achieve better accuracy with a shallow ResNet model. In the TWNN model, the same parameters are trained for a more extended period. Therefore, the corresponding memory requirement is lower than traditional DNN models.
\section{Conclusion}
TWNN neural network leverages fixed-point iteration with automatic differentiation.  In addition, it is capable of explicitly providing more efficient performance by computing gradients for input, hidden states and parameters. As only the influential layers are trained for a longer time, this new model can optimize the hidden dynamics learning. The proposed model reaches accuracy faster than the related Neural Network as each of the batches is training under fixed point constraint through a fixed point iterated block. In the TWNN model, for some batches, the accuracy is achieved faster than other batches.  Therefore, the optimization of the TWNN model is more straightforward in comparison with other Deep Neural Networks.

\bibliographystyle{unsrt}
\bibliography{references}
\end{document}